\documentclass{article} % For LaTeX2e
\usepackage{nips15submit_e,times}
\usepackage{hyperref}
\usepackage{url}

\usepackage{amsmath,amssymb}
\usepackage{algorithm}
\usepackage{algpseudocode}

\usepackage{graphicx}
\usepackage{hyperref}
\usepackage{subfigure} 
\usepackage{wrapfig}
\usepackage{mwe}
\usepackage{multicol}% http://ctan.org/pkg/multicols

\usepackage{mathtools}

\title{Convex Optimization For Non-Convex Problems via  Column Generation  }

\author{
Julian Yarkony\\
Experian Data Lab \\% and Salk Institute$.^0$ \footnotemark[0]\\
San Diego CA \\
\texttt{julian.e.yarkony@gmail.com} \\
\And
Kamalika Chaudhuri \\
University of California San Diego\\
San Diego CA \\
\texttt{kamalika@cs.ucsd.edu}\\
}

\nipsfinalcopy % Uncomment for camera-ready version

\begin{document}

\maketitle

\begin{abstract} 
We apply column generation to approximating complex structured objects via a set of primitive structured objects under either the cross entropy or L2 loss. We use L1 regularization to encourage the use of few structured primitive objects.  We attack approximation using convex optimization over an infinite number of variables each corresponding to a primitive structured object that are generated on demand by easy inference in the Lagrangian dual.  We apply our approach to producing low rank approximations to large 3-way tensors.  
\end{abstract}

\section{Introduction}

We consider the general problem of approximating a complex structured object  using a non-negative weighted combination of primitive structured objects under regularization \cite{terrycite,LDACITE,baldi2012autoencoders}.  Given a class of primitive structured objects (primitives) the set of complex objects is any weighted combination of the primitives or any weighted combination lying in a convex hull or cone\cite{boyd}.  Such problems are common throughout the machine learning literature and include, sparse coding \cite{sparsecode}, matrix approximation \cite{Roweis98emalgorithms,eigenfaces}, tensor approximation \cite{tensorsurvay}, mixture modeling, autoencoder modeling \cite{baldi2012autoencoders}, etc. For example in 3-way tensor (three dimensional) approximation \cite{animatensor,commdetect,tensorbook,anandkumar2012method} the set of rank-1, 3-way tensors are the primitives; the positive cone over rank-1 3-way  tensors is the space optimized over.%; and a triple co-occurrence tensor constructed on real data is the target.  

 In this paper we study two families of complex structured object approximation problems.  The first (Family One) considers minimizing the L2 loss between a complex structured object versus its approximation with L1 regularization \cite{lars} encouraging the use of fewer primitives in the approximation.  One example of a problem under Family One is lossy compressing an image so that the colors in the uncompressed image are similar in an L2 sense to the original \cite{eigenfaces}. % via  L1 regularization.%This  is  L0 regularization over an L2 loss function. However the  L0 loss function is difficult to treat use the convex L1 regularization instead.    
 
 The second family (Family Two) considers minimizing the cross entropy loss between a complex structured object corresponding to a probability distribution versus its approximation with L1 regularization encouraging a fewer number of components in the approximation.  Sparsity in the L1 sense is achieved by encouraging more of the distribution to be explained by the uniform distribution or other white noise, or high entropy distribution.  One example of a problem under Family Two is fitting a mixture of gaussians to describe a probability distribution.   Here the high entropy distribution could be the maximum likelihood gaussian fit to all of the data.   
 
 The two families differ in the difficulty of approximation. Approximation in Family One is easier than approximation in Family Two, and thus Family One can be applied in larger scale applications.  However the cross entropy loss of Family Two is a far more appropriate loss function in domains where probability distributions are being studied, which is common in machine learning. %explicit probability distributions are desired as output such as in: mixture models, topic models, hidden Markov modes or tensors of moments of variables. 
 
 A key difficulty in  complex structured object approximation is that the number of primitive structured objects may be infinite or at least exponential and prevent the construction of a quality approximation.  Much previous work in machine learning relies on non-convex methods such as Expectation Maximization (EM)\cite{Dempster77maximumlikelihood,gaussmix,baldibook,wellingunsupold} which make greedy local moves to improve their approximation. Gradient descent also has great practical value especially in the domain of deep neural networks \cite{hinton,baldi2014searching,pollastri2002improving,krizhevsky2012imagenet}% This can be understood as fixing the value of the L0 norm then 
 
 An alternative approach to solving optimization problems is   column generation \cite{cuttingstock,barnprice,HPlanarCC}.  Column generation is a powerful method generally used for massive scale integer programming problems originating in operations research such routing flights for large airlines.  In this line of work one optimizes over the entire space of exponential number of variables in continuous space.  The corresponding LP relaxation is a convex optimization and given a finite number of variables can be solved via interior point methods\cite{boyd,boydl1} in polynomial time.  However the set of all variables can not be enumerated much less used in optimization.  Column generation operates by solving the optimization problem given a small subset of the variables.  Next one or more variables are identified that will improve the objective if added to the subset under consideration and these are added to the subset under consideration.  The optimization problem is then resolved. This continues until convergence.  %Branch and price methods are used to ensure integral solutions are produced though these are more costly \cite{barnprice}. 
  
  In many optimization problems analysis of Lagrange multipliers allows for the exact or near exact computation of the best variable to add to the subset under consideration.  Such analysis consists of solving a dramatically simpler version of the original problem which is often polynomial time solvable, or at least can be approximated to high accuracy in polynomial time perhaps with guarantees.  In this paper we apply column generation to complex structure approximation in such a way as to circumvent many concerns pertaining to local optima.  To our knowledge we are the first to apply column generation to structured object approximation.  
 
%In this document we present a general treatment of the two families problems of problems above with a novel solution inspired by  column generation.

\subsection{Outline} 
We now provide an outline of this paper.   In Section \ref{formal} we formally introduce our two families of problems.  Next in Section \ref{formopt} we formulate the problems in terms of  convex optimization over enormous or possibly infinite spaces of variables each corresponding to a primitive.  Then in Section \ref{generalInf} we formulate optimization in the form of column generation where variables are generated as needed so as to make convex optimization feasible.  In Section \ref{Violater} we show how to identify primitives to add to consideration to improve the objective.  We show examples derivations in the domain of tensors (Sections \ref{tensec1},\ref{tensec2}) and gaussian mixture models (Section \ref{gsec}).

  In Section \ref{experiment} we show experimental results on tensor problems. We show as a function of time and iteration the value of results from optimization for various size synthetic symmetric 3-way tensor problems.  We use examples from both Family One and Two.  In Section \ref{litReview} we briefly discuss relevant papers and their relationship to our work.  In Section \ref{conc} we discuss future work and extensions. 
 \section{Formal Model}
 \label{formal}
 
 We now formally discuss Family One and Two.  Consider a structured object consisting of a finite number of values denoted $\bar{T}$.  Here $\bar{T}$ may correspond to a tensor or a probability distribution for example.  We denote the vectorization of $\bar{T}$  by column vector $T$.  We denote the set of primitives that may be used to construct $T$ as $\mathcal{M}$ and reference its members with $m$.  Primitives may be rank-1 tensors or gaussian distributions in the previous examples.  We define a matrix $M$ by horizontally stacking the column vectors $m \in \mathcal{M}$ together.  We define a non-negative weighted combination of the columns of $M$ using a non-negative column vector $w$ of length equal to $|\mathcal{M}|$.  

In this document we study the two families in parallel since the approaches for approximation introduced in this paper are highly related.  Also much of the notation used to discuss the two families is shared.  Similarly many problems that are tackled using Family One have corresponding problems in Family Two.  

 \subsection{Family One}
 We  define the optimal model  according to L2 loss under  L1 regularization as follows.  We use $t$ to denote transpose.  We use $\vec{\ell}$  to define a positive constant column vector of length $|\mathcal{M}|$ where the constant value is $\ell \in \mathcal{R}_{+}$. 
 
 \begin{align}
 \label{fam1}
 \min_{w \geq 0} \frac{1}{2} (T-Mw)^t(T-Mw)+\vec{\ell}^tw
 \end{align}
  
   \subsection{Family Two}
 We now define the optimal model  according to cross entropy loss under  L1 regularization as follows.  We use $\log$ to denote the element-wise logarithm.  We use $\vec{\ell}_a$ to define a positive constant column vector of length $|\mathcal{M}|$ where the constant value is $\ell_a \in \mathcal{R}_+$.  In addition $\vec{\ell}_a$  that has a single zero valued entry.  This entry is associated with a special column $m_0$ which corresponds to the uniform distribution or other white noise distribution.  The entry of $w$ corresponding to $m_0$ is denoted $w_0$.  This is useful in modeling probability distributions where $T$ is a complex probability distribution and each $m$ is a primitive probability distribution.  The corresponding optimization problem is written below.  
  \begin{equation}
  \label{fam2}
 \min_{\substack{w \geq 0 \\ w^t 1 =1}} -T^t\log (Mw)+\vec{\ell}_a^tw
 \end{equation}
 Since the Eq \ref{fam2} is non-decreasing in $w_0$ we write Eq \ref{fam2} as follows.  
   \begin{equation}
  \label{fam2a}
\mbox{Eq } \ref{fam2}= \min_{\substack{w \geq 0 \\ w^t 1  \leq 1}}- T^t\log (Mw)+\vec{\ell}_a^tw
 \end{equation}
 
  \section{Formulating Optimization}
   \label{formopt}

  We now consider the treatment of structured object approximation using the tools of quadratic programming (QP) for Family One and linear programming (LP) for Family Two.  
  
\subsection{Family One}
 We now introduce a surrogate object $K$ which corresponds to element-wise distance between $T$ and $Mw$.  We now write optimization as a quadratic program.  
  \begin{align}
  \label{fam2primalbef}
\mbox{Eq } \ref{fam1} = \min_{\substack{w \geq 0 \\ K \geq 0}} \frac{1}{2}K^t K+\vec{\ell}^tw\\
\nonumber \mbox{s.t. }K \geq T-Mw\\
\nonumber K \geq Mw-T
 \end{align}
Here Eq \ref{fam2primalbef} corresponds to a convex quadratic program with an intractable $|\mathcal{M}|$ number of variables (one for each primitive) and a small number of constraints equal to $2|T|$.  We refer to Eq \ref{fam2primalbef} as the primal problem for Family One and it is associated with a dual problem, which is also a convex quadratic program.  This dual problem is written below using dual variables   $\psi,\lambda$ each of which are of length equal to $|T|$.  
 
 \begin{align}
\mbox{Eq }\ref{fam1}=\max_{\substack{\lambda \geq 0\\ \psi \geq 0}} \frac{-\lambda^t\lambda}{2} -\psi^t\lambda -\frac{\psi^t\psi}{2}+T^t \lambda- T^t\psi\\
\nonumber \vec{\ell} \geq M^t\lambda-M^t \psi
\end{align}
 The dual problem above is derived in  Appendix \ref {fam1dualderiv}.

\subsection{Family Two}
 We now introduce a surrogate object $K$ which corresponds to the element-wise logarithm of $Mw$.
  \begin{align}
  \label{fam2primal}
\mbox{Eq } \ref{fam2} = \min_{\substack{w \geq 0 \\ 1^t w\leq 1\\ K \geq 0}} T^tK+\vec{\ell}_a^tw\\
\nonumber \mbox{s.t. }-K \leq  \log(Mw)
 \end{align}

Since the $\log$ function is concave we express it as  the lower envelope of a set of affine upper bounds each of which is constructed via a first order Taylor expansion.  We write this formally below.  %the min of set of upper bounds each constructed from the local taylor expansion of $\log$.  
\begin{align}
\label{logupper}
\log(y)=\min_{\eta \in (0,\infty)}\log(\eta)+(y-\eta)\frac{1}{\eta} \quad  \forall y \in (0,\infty)
\end{align}

We now apply the lower envelope expression of $\log$ in Eq \ref{logupper} to Eq \ref{fam2}. This produces the linear program below.%with this construction of $\log$. 

  \begin{align}
  \label{fam2expand}
\mbox{Eq } \ref{fam2} = \min_{\substack{w \geq 0 \\ 1^t w \leq 1\\ K \geq 0}} T^tK+\vec{\ell}_a^tw\\
\nonumber \mbox{s.t. }-K \leq  \log(1 \eta) +\frac{(Mw-1\eta)}{\eta} \quad  \forall \eta \in (0,1]
 \end{align}
 
 We refer to  Eq \ref{fam2expand} as the primal problem for Family Two.  The dual form  of Eq \ref{fam2expand} which is also a linear program, is written below using dual variables $\alpha$ and $\beta$. Here  $\alpha$ is a scalar and $\beta$ is associated with a unique vector of cardinality $|T|$ for every $\eta$.  % a vector of Lagrange multipliers for every unique $\eta$ denoted $\beta^{\eta}$
 
 \begin{align}
\max_{\substack{\alpha \geq 0\\ \beta \geq 0}} -\alpha +\sum_{\eta \in (0,1]}(1-1\log(\eta))^t\beta^{\eta}\\
\nonumber T \geq \sum_{\eta \in (0,1]}\beta^{\eta}\\
\nonumber \vec{\ell}_a+1\alpha \geq M^{t} \sum_{\eta \in (0,1]} \frac{1}{\eta} \beta^{\eta}
\end{align}

 The dual problem above is derived in  Appendix \ref {fam2dualderiv}.

 \section{Inference in General Terms}
\label{generalInf}
Solving a quadratic or linear program is done by various methods such as interior points or simplex.  However to employ them there must be a finite  number of variables and constraints and our problems do not satisfy this criteria.  To circumvent it we solve altered versions of the problems that consider only a subset of the constraints and variables. Variables or constraints are then added when and if they are violated or would improve the objective. We iterate between solving the altered problems and adding variables or constraints. The dual problems facilitate the addition of primal variables via analysis of the dual  variables. % At any time a live solution can be produced by the primal solution.% Many of those methods provide primal values and dual values. 

 \subsection{Family One}
 
We solve problems in Family One by solving the dual problem.  In order to solve the dual problem we must identify a subset of the constraints such that when  enforced no other constraints are violated.  We build that subset denoted $\hat{M}$, which is called the working set, greedily. Here $\hat{M}$ is initialized to the empty set or with any subset of the columns of $M$.  We write the corresponding quadratic program below.  % solve the following QP

 \begin{align}
 \label{duallp2}
\mbox{Eq }\ref{fam1}\leq \max_{\substack{\lambda \geq 0\\ \psi \geq 0}} \frac{-\lambda^t\lambda}{2} -\psi^t\lambda -\frac{\psi^t\psi}{2}+T^t \lambda- T^t\psi\\
\nonumber \vec{\ell}\geq \hat{M}^t\lambda-\hat{M}^t \psi
\end{align}

 Finding the most violated constraint corresponds to selecting a column of $M$ to maximize the following for any $\lambda$ and $\psi$.  For ease of notation we define a term $\theta=\lambda-\psi$.
 
 \begin{align}
 \label{maxm1}
 \max_{m \in \mathcal{M}}m^t(\lambda-\psi^t)= \max_{m \in \mathcal{M}}\theta^t m
 \end{align} 
 
 Consider a setting where we are able to identify the maximizing argument  $m$ for Eq \ref{maxm1}.  The corresponding column generation algorithm is written in Alg \ref{dualsolvefam1}.   
 
\begin{algorithm}
\caption{Dual Optimization }
\begin{algorithmic} 
\State $\hat{M} \leftarrow \{ \}$  \\
\While{True}
\State $[\lambda,\psi] \leftarrow$ Solve Eq \ref{duallp2}  given $\hat{M}$ % \ref{DualLpFinProperA} given $
\State $\theta \leftarrow \lambda- \psi$% \ref{DualLpFinProperA} given $
\State $[m] \leftarrow \max_{m\in \mathcal{M}} \theta^t m$
\If{ $\theta^t m > \ell $}
\State $\hat{M} \leftarrow [\hat{M},m]$
\Else 
\State BREAK
\EndIf
 \EndWhile
\end{algorithmic}
\label{dualsolvefam1}
\end{algorithm}

At the termination of Alg \ref{dualsolvefam1} it is the case that Eq \ref{duallp2}= Eq \ref{fam1}.  Also at any time the primal variables $w$ which are produced concurrently with $\lambda,\psi$ describe a valid sub-optimal solution to Eq \ref{fam1} when using interior points methods.  At termination additional sparsity can be created by running least angle regression given the basis $\hat{M}$ \cite{lars}.%and this will improve with each iteration.  
 
\subsection{Family Two} 

Applying column generation to Family Two is  challenging because there are an infinite number of variables in both the primal and dual.  To meet this difficulty we store a subset of the constraints/variables  to the primal and  dual.  We then produce a solution using only those constraints/variables.  We then identify those violated constraints and add them to the constraint set.  Constraints added in the primal correspond to new variables in the dual and similarly constraints added in the dual correspond to new variables in the primal.  

To assist our discussion we index $T$ as follows.  We index $T$ with $x$ where $T_x$ is  the $x$'th value of vector $T$.  We index $K$ and $Mw$ similarly.  We use $\mathcal{\hat{S}}$ to denote the working subset of $x,\eta$ pairs and use $\hat{M}$ to denote the working subset of $M$. We initialize $\hat{M}$ with the uniform distribution/white noise/high entropy distribution $m_0$ and initialize $\mathcal{\hat{S}}$ to include $[x,\frac{1}{|T|}]$ for all $x$.  The initial setting of $\hat{S}$ is an arbitrary choice that worked well in practice.  We use $m$ to index the columns of $M$.  The corresponding primal and dual pair are below.  %
  \begin{align}
  \label{p1p}
 \min_{\substack{w \geq 0 \\ 1^t w\leq 1 \\ K \geq 0}} T^tK+\vec{\ell}_a^tw\\
\nonumber \mbox{s.t. }-K_x \leq  \log(\eta) +\frac{1}{\eta }(\hat{M}w)_x-1 \quad  \forall [x,\eta]\in \mathcal{\hat{S}}
 \end{align}
 
\begin{align}
  \label{p1d}
\mbox{Eq } \ref {p1p} = \max_{\substack{\alpha \geq 0\\ \beta \geq 0}} -\alpha +\sum_{x,\eta \in \mathcal{\hat{S}}}(1-\log(\eta))\beta^{\eta}_{x}\\
\nonumber T_x \geq \sum_{\substack{[\dot{x},\eta] \in \hat{S}\\s.t. \dot{x}=x}}\beta^{\eta}_x \quad \forall x  \\
\nonumber \vec{\ell}_a+1\alpha \geq \sum_{x,\eta \in \hat{S}} m_x\frac{1}{\eta} \beta^{\eta}_x \quad \forall m \in \hat{M}
\end{align}

For ease of notation we  introduce a vector $\theta$ defined as follows.
\begin{align}
\theta_x=\sum_{\substack{[\dot{x},\eta] \in \hat{S}\\s.t. \dot{x}=x}}\frac{1}{\eta} \beta^{\eta}_x \quad \forall x
\end{align}
Consider that we are able to identify violated constraints in the primal and dual.  We then apply the following iteration.  We solve Eq \ref{p1d} which provides us with a solution to  Eq \ref{p1p} as well as Eq \ref{p1d}.  We then identify violated constraints in the primal and dual.  We then add those to the working sets $\hat{\mathcal{S}}$ and $\hat{M}$.

Finding constraints that are violated in the primal is trivial.  For each $K_{x}$ we  minimize  with respect to $\eta$ the following  $K_x+\frac{1}{\eta}(\hat{M}w)_{x}+\log(\eta)-1$.  We take the derivative with respect to $\eta$ then set the derivative equal to zero and finally solve for $\eta$.  
\begin{align}
0=\frac{-1}{\eta^2}(\hat{M}w)_{x}+\frac{1}{\eta}  \\
\nonumber (\hat{M}w)_x=\eta
\end{align}
We then add the pair $x,\eta$ to $\hat{S}$ if the constraint is violated.  We write the corresponding optimization algorithm for Family Two in Alg \ref{dualsolvesimple}.  
\begin{algorithm}
\caption{Dual Optimization }
\begin{algorithmic} 
\State $\hat{M} \leftarrow \{m_0 \}$  \\
\State $\hat{S} \leftarrow \{x,\frac{1}{|T|} \}$ $\forall x$ \\
\While {True}
\State $[\alpha,\beta,w,K] \leftarrow$ Solve Eq \ref{p1p}/\ref{p1d}  given $\hat{M}$ and $\hat{\mathcal{S}}$ % 
\For {x}
\State  $\hat{\mathcal{S}}\leftarrow \hat{\mathcal{S}} \cup (x,(\hat{M}w)_x)$
\EndFor
\State $[m] \leftarrow \max_{m\in \mathcal{M}}   m^t\theta $
\If{ $\theta^t m > \ell_a+\alpha $ }
\State $\hat{M} \leftarrow [\hat{M},m]$
\EndIf
\If {No constraints added this round}
\State BREAK
\EndIf
 \EndWhile
\end{algorithmic}
\label{dualsolvesimple}
\end{algorithm}
 As in optimization in Family One, additional sparsity can be created by running least angle regression given $\hat{M}$ \cite{lars}.

\section{Identifying the Most Violating $m$ or Highly Violated $m$}
\label{Violater}
The previous sections reference solving for the optimal $m$ as the following optimization  
$\max_{m \in M}\theta^t m$.   The difficulty of solving for the optimal $m$ is problem and problem instance specific.  However these problems tend to relatively easy.

Consider the problem of approximating a high rank 3-way tensor with a low rank tensor under  L2 or cross entropy loss and L1 regularization.  In this section we demonstrate that solving for $m$ fits the optimal rank-1 tensor to a tensor described by the dual variables.  %This is desirable because To provide intuiIt can  which is a  far easier than fitting a rank two tensor.  

In the case of gaussian mixture models we demonstrate that solving for $\max_{m \in \mathcal{M}}\theta^t m$ corresponds to training the maximum likelihood gaussian under a weighting of the points described by $\theta$ which can be solved exactly in closed form.  In fact $\theta$ is an unnormalized probability distribution.  Here and in all Family Two examples $\theta$  must be normalized before applying inference and shifting the scaler outside of the $\max$ in $\max_{m \in \mathcal{M}}\theta^t m$.  Normalization is done by dividing each element of $\theta$ by $1^t \theta$.   %during optimization to aid in understanding.    

It should be observed that local optima are not very problematic when identifying $m$.  One simply needs to find a local optima that has objective value greater than $\ell$ or $\ell_a+\alpha$ should one exist. This results in Alg \ref{dualsolvefam1}/ \ref{dualsolvesimple} changing the dual variables and hence $\theta$ and allow for one to try again to find a good local optimum on a different and perhaps easier problem.  Finding a poor local optimum never results in an increase the objective for either Family One or Two.  If a poor quality $m$ is included early in the column generation process it is given zero weight $w$ by the optimizer.  Multiple local optima each corresponding to a violated constraint can be computed here and added to $\hat{M}$.  

We consider three examples of finding the most violated $m$ below and with additional examples in Appendix \ref{nonsym1},\ref{nonsym2}.  

\subsection{Example, Family One:  Symmetric 3-way Tensors}

\label{tensec1}
We now study $\max_{m \in M}\theta^t m$  in the domain where $m$ corresponds to fitting a symmetric 3-way tensor.  Consider the case that $T$ corresponds to the vectorization of a symmetric 3-way tensor.  Thus each column of $M$ also corresponds to a symmetric 3-way tensor.  We describe $m$ using a vector $v$.  Here $v \in \mathcal{R}^{(|T|^{\frac{1}{3}})}$ where $v$ is a unit vector.  The non-vectorized form of $m$ is denoted $\bar{m}$ and is indexed by $i,j,k \in \{0,1,2...N-1\}$.  We define $\bar{m}$ in terms of $v$ below.

\begin{align}
\bar{m}_{ijk}=v_iv_jv_k
\end{align}

We write optimization below. 
%.  Thus $v\geq 0$ and $1^tv = 1$
\begin{align}
\max_{m \in M}\theta^t m=\max_{\substack{v \\ v^tv = 1} }\sum_{ijk}\theta_{ijk}v_i v_j v_k
\end{align}

The projected gradient update for $v_i$ for all $i$ is written below.

\begin{align}
\dot{v}_i \leftarrow v_i+(\mbox{stepsize}) \cdot \sum_{jk}\theta_{ijk}v_j v_k 	\quad \forall i\\
\nonumber v \leftarrow \frac{\dot{v}}{\dot{v}^t\dot{v}}
\end{align}

Remarkably the seminal power iteration \cite{mises1929praktische,pagerank} can be applied in place of projected gradient descent.   Furthermore for tensors convergence of the  power iteration to a local optima of the objective can be guaranteed \cite{animatensor}.  Unlike in the case of 2-way tensor (matrix) global optimality is not guaranteed.  The corresponding updates are written below.   
\begin{align}
\dot{v}_i \leftarrow \sum_{jk}\theta_{ijk}v_j v_k\\
\nonumber v\leftarrow \frac{\dot{v}}{\dot{v}^t\dot{v}}
\end{align}
We repeatedly update $v$ until convergence.  At termination we add $m$ and $-m$ to the working set $\hat{M}$ when applying the power iteration.  This is because the power iteration maximizes the magnitude of $\theta^tm$ without concern for the sign.   

\subsubsection{Note on Even way Tensors and Optimization}
For even way tensors (2-way,4-way,6-way...)  an outer product does not produce all possible rank-1 tensors for $M$.  It fails to create those constructed by an outer product then multiplied by $-1$.  This can be seen by observing the following: multiplying the vector $v$ by $-1$ does not flip the sign of all elements of $m$ (it flips none) while this is achieved for odd way tensors (3-way, 5-way,7-way...). Thus at termination of optimization power iteration optimization we add $m$ and $-m$ to the working set $\hat{M}$.

 Projected gradient optimization must be similarly altered in the case of even way tensors.   This is done by computing $\max_{m \in M}\theta^t m$ and $\max_{m \in M}(-\theta)^t m$ and adding $m$,$-m$ to $\hat{M}$ corresponding to violated constraints.
 
 This is important because Family One is restricted to have non-negative components $w$ to fit in the standard form for quadratic programming however there need not be a model constraint that the weights are non-negative.  

 % Even way tensors must be treated in this manner because multiplying the vector $v$ by $-1$ does not flip the sign of all elements of $m$ (it flips none) while this is achieved for odd way $v$.  Thus for even way tensors outer an product does not achieve all possible elements of $M$ unless an external sign is applied.

\subsection{Example Family Two:  Fitting a symmetric 3-way tensor defined by  probability distribution}  

\label{tensec2}

We now study $\max_{m \in M}\theta^t m$  in the domain where $m$ corresponds to fitting a symmetric 3-way tensor in the setting of Family Two.  We use the notation of Section \ref{tensec1}.  %.  Consider the case that $T$ corresponds to the vectorization of a symmetric 3-way tensor.  Thus each column of $M$ also corresponds to a symmetric 3-way tensor.  We use describe $m$ using a vector $v$.  Here $v \in R^3$ where $v$ is a unit vector.  The non-vectorized form of $m$ is denoted $\bar{m}$ and is indexed by $i,j,k \in \{0,1,2...N-1\}$.  We define $\bar{m}$ in terms of $v$ below.
In this section we solve optimization relying on Jenson's inequality \cite{jensen1906fonctions, jordonbook} to force $m$ to correspond to a probability distribution at all times and avoid gradient descent. We write optimization below.  
 \begin{align}
 \max_{\substack{v\geq 0 \\ 1^tv = 1}} \sum_{ijk}\theta_{ijk}v_i v_j v_k
 \end{align}
 Recall that  $\theta$ is non-negative and we normalize $\theta$ to sum to one and shift the normalization constant outside of the $\max$. We now write optimization over the $\log$ of $\max_{m \in M}\theta^t m$.
  \begin{align}
  \label{befz}
 \max_{\substack{v\geq 0 \\ 1^tv = 1}} \log(\sum_{ijk} \theta_{ijk}v_i v_j v_k)
 \end{align}
 
 As in EM methods for inference in probabilistic models we define a proposal probability distribution $z$ indexed by $ijk$.  We initialize $z$ as follows to reflect the probability distribution $\theta$ though this initialization heuristic and random initialization of $z$ is also valid. 
 \begin{align}
 z_{ijk}\leftarrow \frac{\theta_{ijk}}{\sum_{\dot{i}\dot{j}\dot{k}}\theta_{\dot{i}\dot{j}\dot{k}}}  %The 6 is to cover the symmetry in $\theta$.  
\end{align}
We now multiply and divide by $z$ as is standard in EM methods and apply Jenson's inequality .  
  \begin{align}
 \mbox{Eq } \ref{befz}=\max_{\substack{v\geq 0 \\ 1^tv = 1}} 
 \log(\sum_{ijk} \frac{z_{ijk}}{z_{ijk}}\theta_{ijk}v_i v_j v_k) \\ %\; \; \forall[ v\geq0; 1^tv=1]\\
\nonumber \geq \max_{\substack{v\geq 0 \\ 1^tv = 1}} \sum_{ijk}-z_{ijk}\log z_{ijk}+z_{ijk}\log \theta_{ijk} \\
\nonumber + \sum_{ijk}z_{ijk}\log v_i+\sum_{ijk}z_{ijk}\log v_j+\sum_{ijk}z_{ijk}\log v_k
 \end{align}

We now add a  Lagrange multiplier $\tau$ to enforce that $v$ sums to one.  There will be no need to enforce non-negativity in optimization.  

  \begin{align}
 \max_{\substack{v\geq 0} }\min_{\tau \in (-\infty,\infty)}\tau(1-1^tv)+\sum_{ijk}-z_{ijk}\log z_{ijk}\\
  \nonumber +\sum_{ijk}z_{ijk}\log \theta_{ijk} +\sum_{ijk}z_{ijk}\log v_i \\
  \nonumber +\sum_{ijk}z_{ijk}\log v_j+\sum_{ijk}z_{ijk}\log v_k
\end{align}

We now write the optimization with respect to $v_i$.  We now take derivative with respect to $v_i$ and set it equal to 0.  Notice that $v_i$ is present in six terms in the derivative.  

\begin{align}
0=-\tau+\sum_{jk} (z_{ijk}+z_{ikj}+z_{jik}+z_{kij}+z_{jki}+z_{kji})\frac{1}{v_i}\\
\nonumber \tau v_i=\sum_{jk} (z_{ijk}+z_{ikj}+z_{jik}+z_{kij}+z_{jki}+z_{kji})\\
 \nonumber v_i=\frac{1}{\tau}\sum_{jk} (z_{ijk}+z_{ikj}+z_{jik}+z_{kij}+z_{jki}+z_{kji})
\end{align}

Observe that $v_i\propto \sum_{jk} z_{ijk}$.  Since it is the case that  $1^t v=1$.  Then the following is true.  
\begin{align}
\tau=\sum_{ijk} z_{ijk}
\end{align}

The optimizing updates for $z_{ijk}$ set $z_{ijk}$ proportional to $\theta_{ijk}v_i v_j v_k$ based on the standard application of the tightest bound for Jenson's inequality.   Therefore the final updates are as follows.  
\begin{align}
\label{zvupdates}
z_{ijk} \propto \theta_{ijk}v_i v_j v_k\\
\nonumber v_i\propto \sum_{jk} z_{ijk}
\end{align}
We repeatedly update $z$ then $v$ until convergence.  
%By combining the updates in Eq \ref{zvupdates} we can produce a set of updates for $v$ ignoring $z$ completely (except for the first iteration).  These are written below and look very similar to the power iteration.  
%\begin{align}
%v^{\mbox{NEW}}_i \leftarrow  \sum_{jk}\frac{\theta_{ijk}v_i v_j v_k}{\sum_{\hat{i},\hat{j},\hat{k}}\theta_{\hat{i}\hat{j}\hat{k}}%v_{\hat{i}} v_{\hat{j}} v_{\hat{k}}}
%\end{align}

\subsection{Example Gaussian Mixture models}
\label{gsec}
Consider that $T$ represents samples drawn from a  continuous probability distribution.  Here $T$ has one index for every one of $N$ samples.  Consider the problem of approximating $T$ using a set of basis functions $\mathcal{M}$.  Specifically consider the case of a gaussian basis on one dimension with fixed variance $\sigma$.  Each $m$ describes a particular gaussian via displaying the density at each data point $x$. 

Let us define each $m$ via a unique  mean $\mu$.  We use $p_x$ to denote the spatial position of a given point $x$. We apply optimization via Jenson's inequality as in Section  \ref{tensec2} and using the corresponding notation.  Consider a probability distribution $z$ indexed by $x$.  We now write the objective  for optimization and apply Jenson's inequality \cite{jordonbook}.  

Recall that  $\theta$ is non-negative and we normalize $\theta$ to sum to one and shift the normalization constant outside of the $\max$. We now write optimization over the $\log$ of $\max_{m \in M}\theta^t m$.

\begin{align}
\max_{m \in M}\log(\theta^tm)=\max_{m \in M}\log \sum_x(\theta_x m_x)\\
\nonumber =\max_{\mu}\log \sum_x\theta_x\frac{1}{\sigma \sqrt{2 \pi}} e^{-\frac{(p_x-\mu)^2}{2\sigma^2}}\\
\nonumber =\max_{\mu}\log( \sum_x\frac{z_x}{z_x}\theta_x \frac{1}{\sigma \sqrt{2 \pi}}e^{-\frac{(p_x-\mu)^2}{2\sigma^2}}) \\
\geq \max_{\mu} \sum_x -z_x\log z_x \\
\nonumber +  z_x \log( \frac{1}{\sigma \sqrt{2 \pi}} \theta_x e^{\frac{(p_x-\mu)^2}{2\sigma^2}} )\quad \forall [z\geq 0,1^tz=1]
\label{bothere}
\end{align}

Given $z$ we can optimize with respect to $\mu$.   Optimizing with respect to $z$ is done as is standard in Jensons' inequality and is written below.  
\begin{align}
z_x \propto \theta_x \frac{1}{\sigma \sqrt{2\pi}}e^{-\frac{(p_x-\mu)^2}{2\sigma^2}}
\end{align}
Similarly given $\mu$ we can optimize $z$.  We now write optimization over $\mu$ given $z$.

\begin{align}
 \max_{\mu} \sum_x -z_x\log z_x \\
 \nonumber +z_x \log( \frac{1}{\sigma \sqrt{2 \pi}}\theta_x e^{\frac{-(p_x-\mu)^2}{2\sigma^2}})\\
\nonumber=\max_{\mu} \sum_x -z_x\log z_x -z_x \log \sigma -z_x\frac{1}{2}\log(2\pi)\\
\nonumber+ z_x \log \theta_x+z_x\frac{-(p_x-\mu)^2}{2\sigma^2}%z_x( \frac{1}{\sigma \sqrt{2 \pi}}\theta_x e^{\frac{(p_x-\mu)^2}{\sigma}})
\end{align}
We now take the derivative with respect to $\mu$ and set it equal to zero and thus we obtain a closed form expression of $\mu$ .  

\begin{align}
0=\frac{2}{2\sigma^2}\sum_xz_x(\mu-p_x)\\
\nonumber \sum_x z_xp_x=\mu \sum_x z_x\\
\nonumber \sum_x z_xp_x=\mu 
\end{align}

We repeatedly update $z$ then $\mu$ until convergence. We begin with a value for $\mu$ then solve for $z$ where we initialize $\mu$ to correspond to a point $p_x$ where $x$ is selected with probability proportionate to $\theta_x$.  This is an initialization heuristic but attempts to place the gaussian in a region of high density of probability mass $\theta$. %and random initialization of .  % Optimizing with respect to $z$ is done as is standard in Jensons' inequality.  
%\begin{align}
%z_x \propto \theta_x e^{-\frac{(p_x-\mu)^2}{2\sigma^2}}
%s\end{align}
%\cite{dadcite,yarkony1974interaction}
\section{Experiments}
\label{experiment}
In this section we show the effectiveness of our approach for approximating large symmetric  3-way tensors in Family One and Two.  
\subsection{Family One}
We now study approximation to symmetric 3-way tensors for Family One to test the effectiveness of our approach.  We use various sizes  (30,35,40,45), constructed from a convex combination of three to ten unique rank-1 tensors with weights summing to one.  We construct each unique rank-1 tensor by the triple outer product of a unique random unit vector.  Each such vector has between six and fifteen non-zero elements. We inject noise describing between one and twenty percent of the tensor.  For each problem instance we use four different L1 regularizers [0.1, 0.4, 0.7, 1.0].   We consider 9000 problems instances and we continue optimization for up to five minutes per instance after which termination is done after solving the current QP.  

  In Fig \ref{quantprb:a1} we show the loss with respect to time averaged over non-terminated problem instances.  This plot demonstrates that we  rapidly produce low cost solutions.

In Fig \ref{quantprb:b1} we show a scatter plot of total optimization time vs the loss at termination where each instance is a single data point.  We show normalized and un-normalized values where normalization corresponds to subtracting the loss on the ground truth model (note that the ground truth model does not model the noise). This plot demonstrates that we are able to fit the tensors rapidly and are able to overfit which is important for an optimization approach. 

In Fig \ref{quantprb:c1} we show the derivative of the objective with respect to the columns in ground truth model basis. This plot demonstrates that little is gained by adding the ground truth basis tensors to $\hat{M}$ if not present at convergence.  From this we conclude that the power iteration is able to find good local optima of $\max_{m \in \mathcal{M}} \theta^tm$.  

%Over 7300 problem instances define these plots (still running computation). 
%We oberve that we are able to fit the data.
\subsection{Family Two}
We test the effectiveness our our approach on symmetric 3-way tensors in Family Two exactly as for Family One though on a smaller scale of problems.  We use tensors of size twenty where the basis vectors used to construct the tensor have between four and six non-zero values.  Each basis vector was non-negative and its elements summed to one instead of having unit norm as for Family One.  We consider 750 problems instances and we continue optimization for up to five minutes per instance after which termination is done after solving the current LP.  

\begin{figure*}
\centering     %%% not \center
\subfigure[]{\label{quantprb:a1}\includegraphics[trim={1cm 6cm 1cm 6cm},clip,width=60mm]{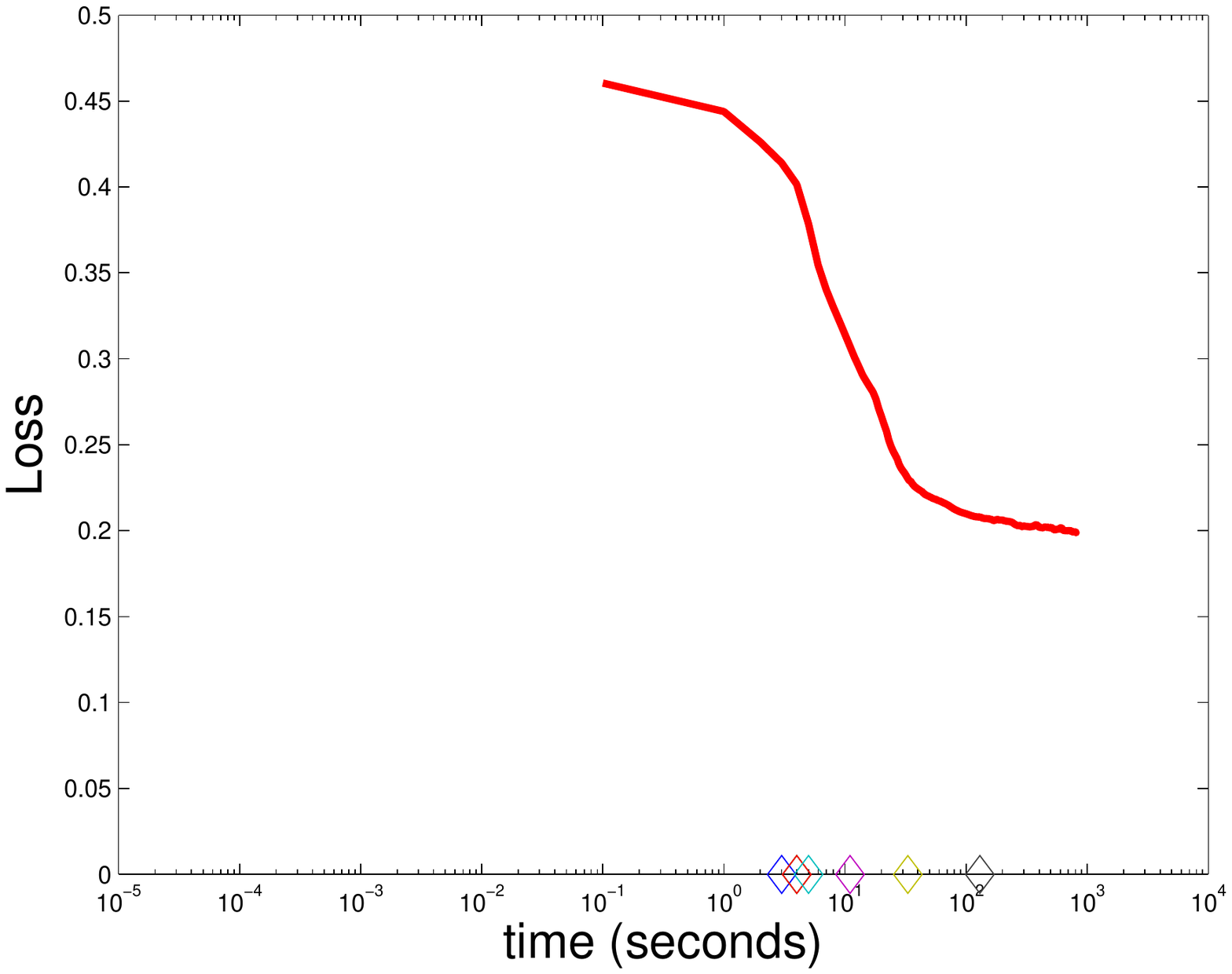}}
\subfigure[]{\label{quantprb:b1}\includegraphics[trim={1cm 6cm 1cm 6cm},clip,width=60mm]{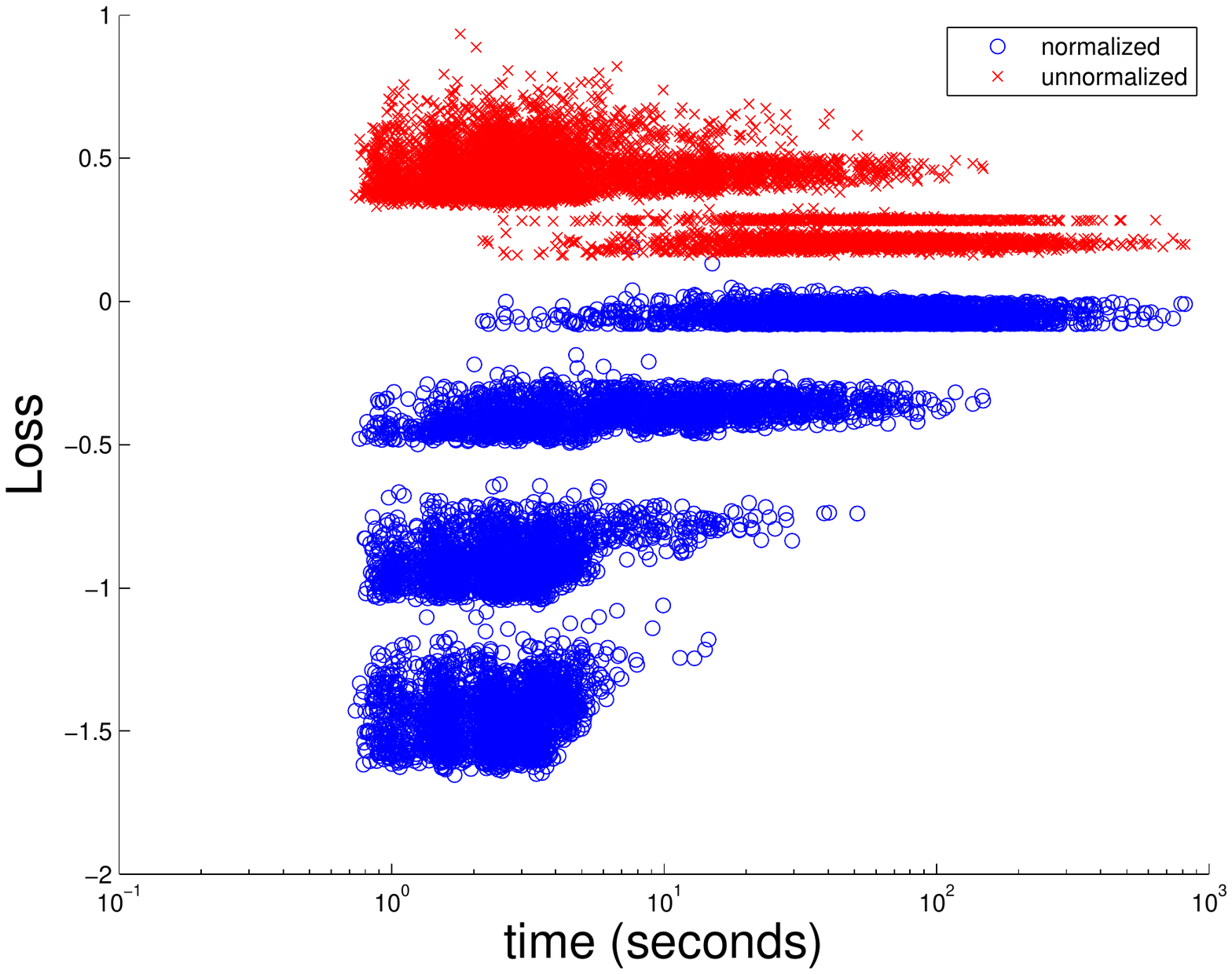}}
\subfigure[]{\label{quantprb:c1}\includegraphics[trim={1cm 6cm 1cm 6cm},clip,width=60mm]{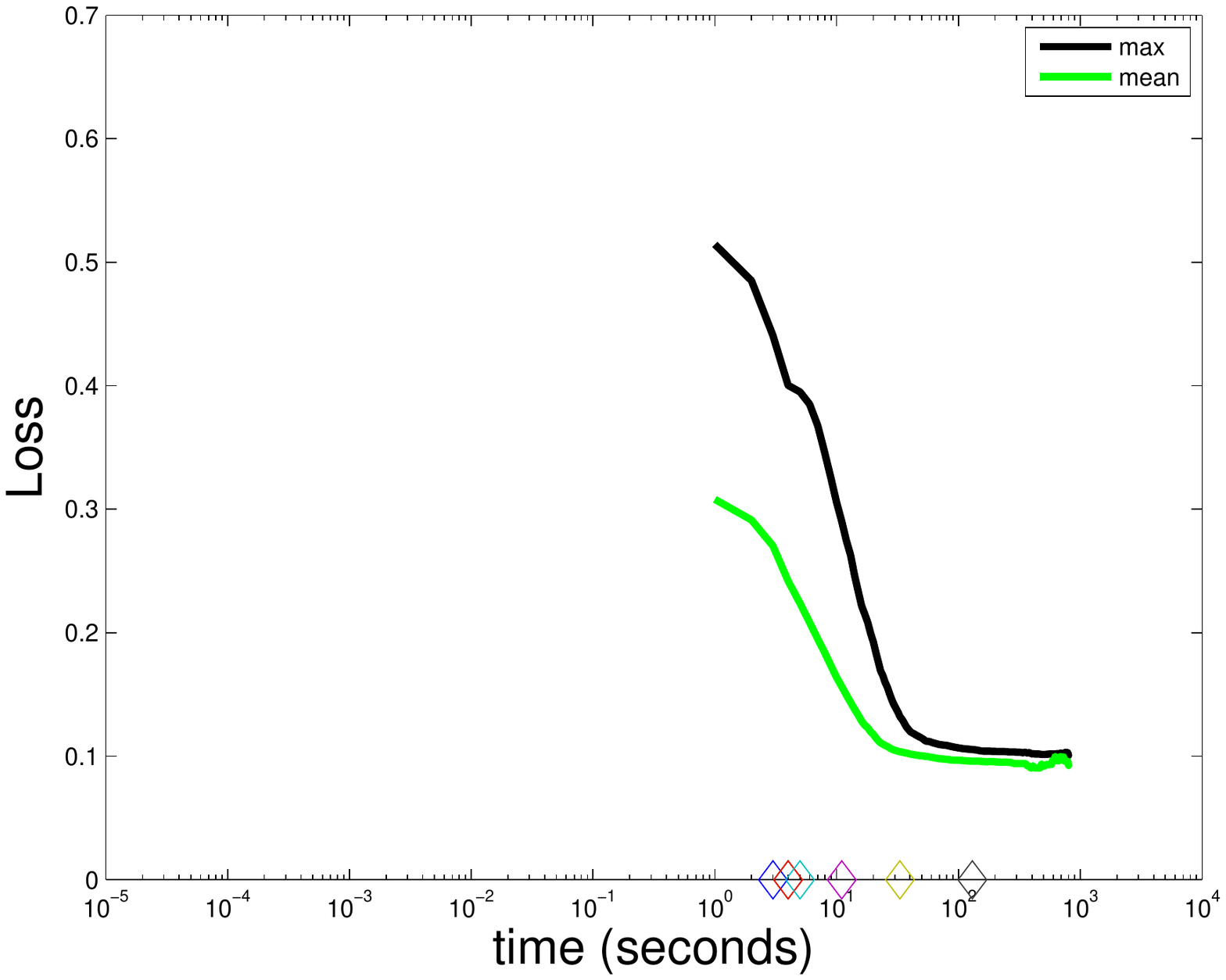}}
\subfigure[]{\label{quantprb:a}\includegraphics[trim={1cm 6cm 1cm 6cm},clip,width=60mm]{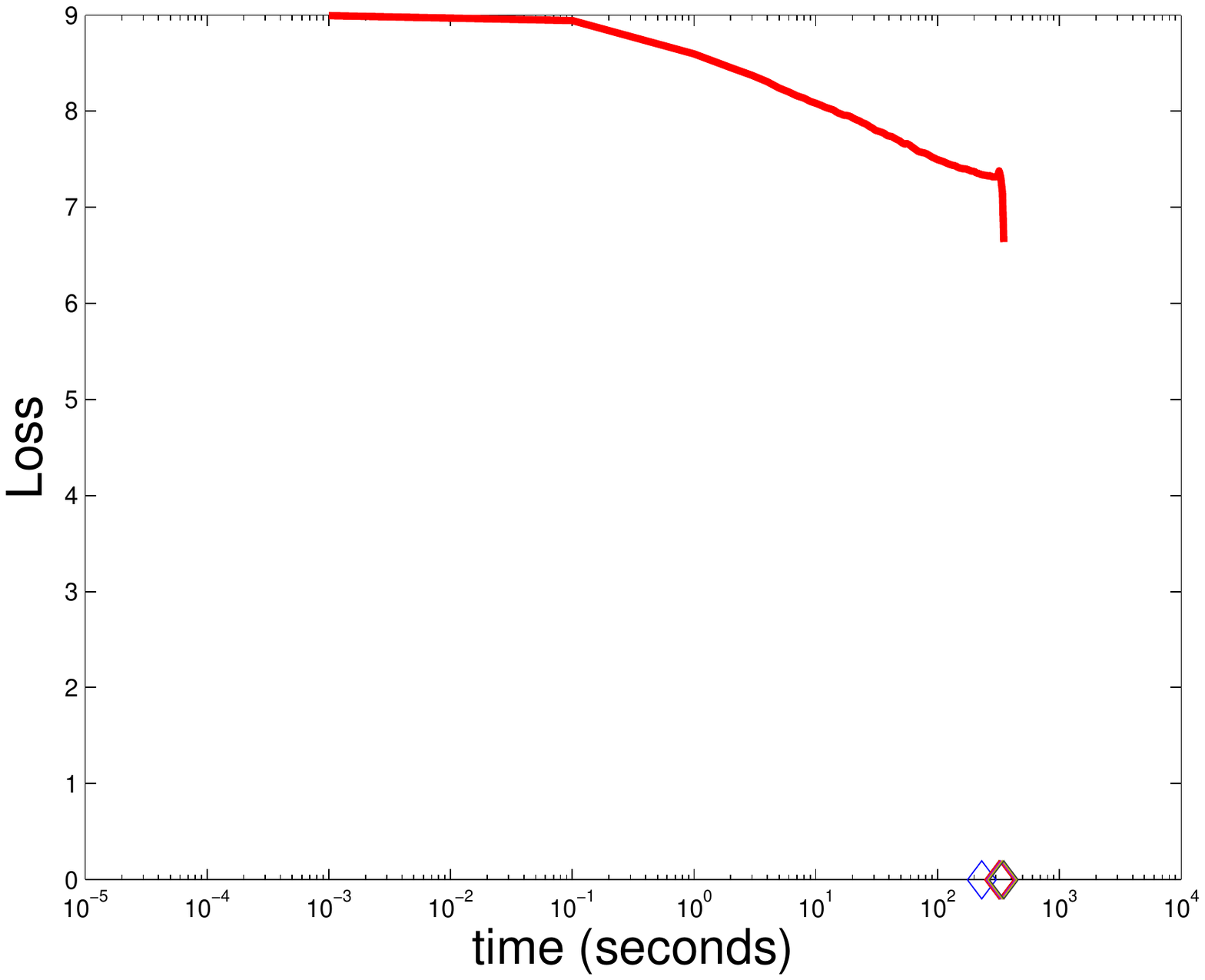}}
\subfigure[]{\label{quantprb:b}\includegraphics[trim={1cm 6cm 1cm 6cm},clip,width=60mm]{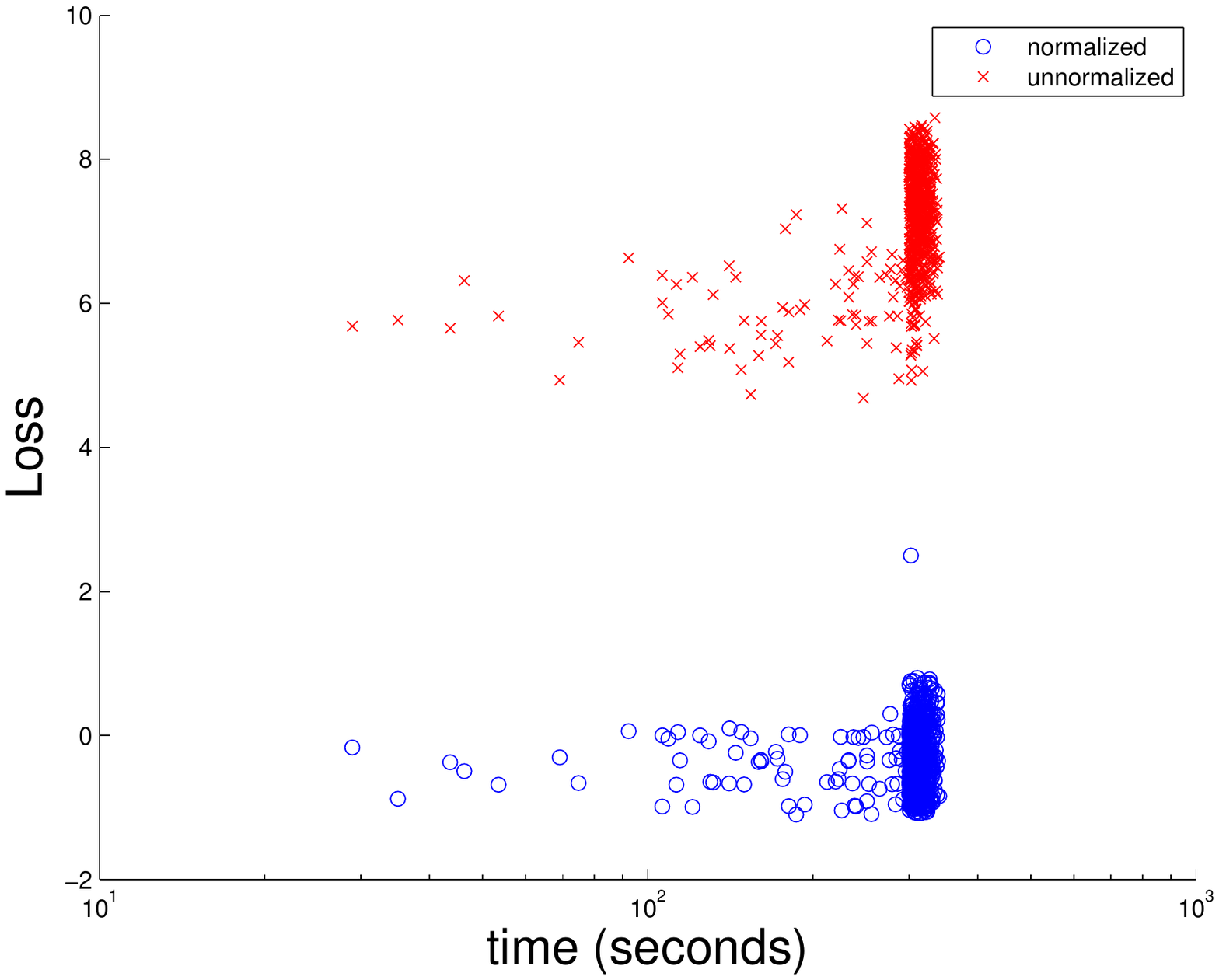}}
\subfigure[]{\label{quantprb:c}\includegraphics[trim={1cm 6cm 1cm 6cm},clip,width=60mm]{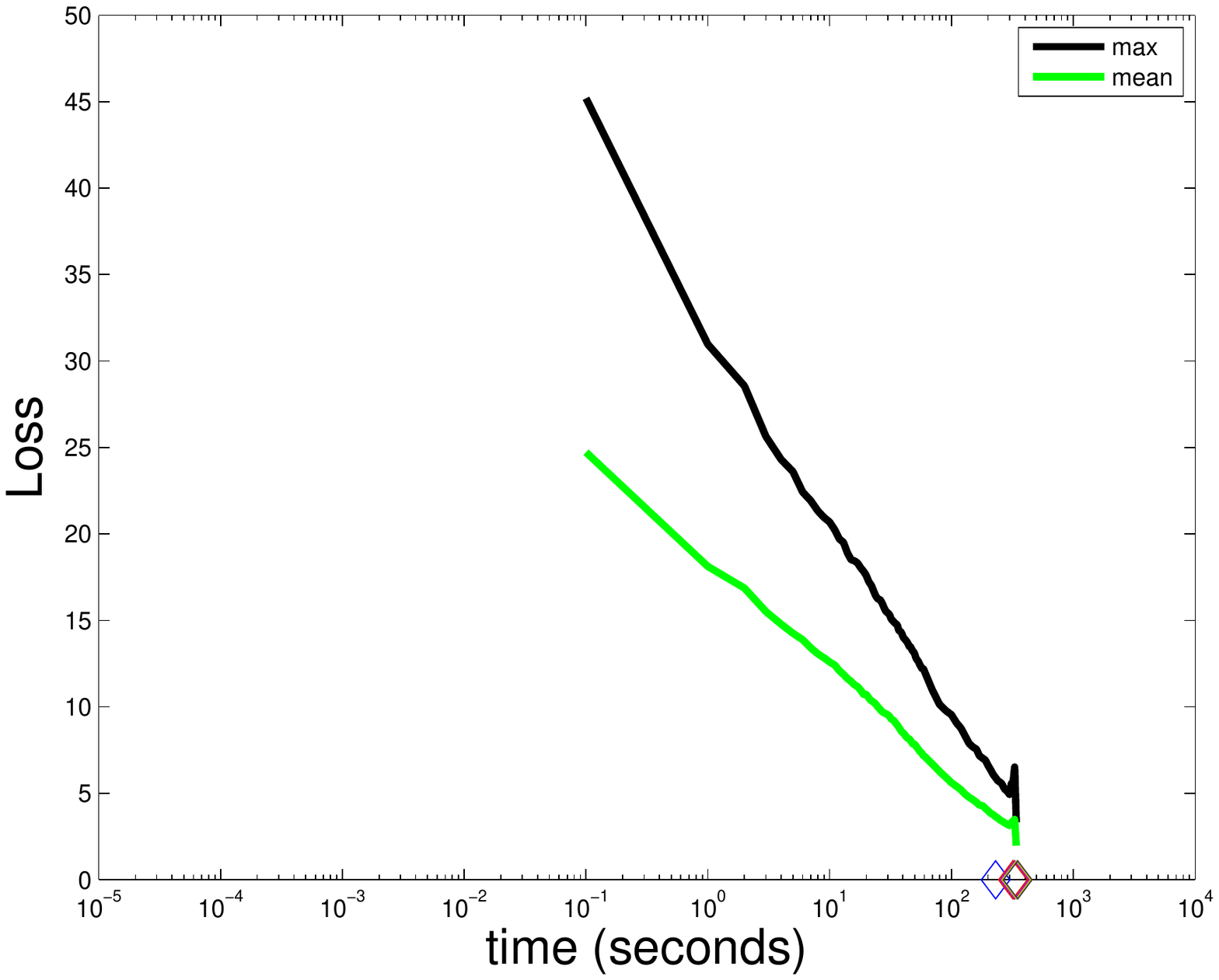}}
%\subfigure[b]{\label{quantprb:b}\includegraphics[width=45mm]{sharinganMultiPlots/sam1}}
%\subfigure[c]{\label{quantprb:c}\includegraphics[width=45mm]{sharinganMultiPlots/sam1}}
\caption{   We show quantities averaged over problem instances.  \ref{quantprb:a1} (Family One) We show loss with respect to time averaged over problem instances that have not terminated. We use diamonds to denote [95,80,65,50,35,20,5] percent not terminated.  \ref{quantprb:b1} (Family One) We show a scatter plot of  total optimization time vs  loss at termination of optimization. Each instance is associated with a pair of points, one associated with normalized loss and other other associated with unnormalized loss.  Normalization corresponds to subtracting the loss corresponding to the model.  \ref{quantprb:c1} (Family One)  We show the derivative of the objective with respect to the columns in ground truth model basis.  We show the maximum and mean value.   We average over non-terminated instances. We use diamonds to denote [95,80,65,50,35,20,5] percent non-terminated.  Fig\ref{quantprb:a},\ref{quantprb:b},\ref{quantprb:c} are the corresponding plots for Family Two for Fig \ref{quantprb:a1},\ref{quantprb:b1},\ref{quantprb:c1} respectively.  } %(3): segmentation with 28 random ground truth long range interactions. (4):  the random pairs of points used  to produce (3) are denoted by displaying their connections. (5,6),, (7,8) are teh same ase (3,4) except with 58 and 508 edges.  } %(5): segmentation with 58 random ground truth long range interactions. (6):  the random pairs of points used  to produce (5) are denoted by displaying their connections.  (7): segmentation with 28 random ground truth long range interactions. (8):  the random pairs of points used  to produce (7) are denoted by displaying their connections.  }
\label{quantprb}
\end{figure*}

 In Fig \ref{quantprb:a} we show the loss with respect to time averaged over non-terminated problem instances.  This plot demonstrates that we  rapidly   produce low cost solutions though not as quickly as in Family One. 

In Fig \ref{quantprb:b} we show the corresponding plot for Family Two Tensors for \ref{quantprb:b1}.  When normalizing we require the objective of the ground truth model.  We add a small amount of probability mass (0.00001) to each entry of the ground truth model so that it does not have any zero values during cross entropy computation.  Thus the sum of the values elements in the ground truth model is greater than one. This plot demonstrates that we are able to fit the tensors rapidly and are able to overfit which is important for an optimization approach. 

In Fig \ref{quantprb:c} we show the corresponding plot for Family Two Tensors for \ref{quantprb:c1}.  This plot demonstrates that little is gained by adding the ground truth basis tensors to $\hat{M}$ if not present at convergence.  From this we conclude that the our Jenson's inequality based optimization is able to find good local optima of $\max_{m \in \mathcal{M}} \theta^tm$.

% (still running computation).  
%\subsection{Family Two:  Gaussian Experiments}
%Zhixuan is attacking this.  I have code to do synthetic gaussians.  

\section{Literature Review}
\label{litReview}
Our work can be positioned at the intersection of two well studied areas: 1) column generation for combinatorial optimization and 2)  efficient representation of data notably with regards to low rank tensor decompositions.  In this section we discuss some of the  related work in these two vast areas in the context of our work.  

\subsection{Column Generation}
\subsubsection{The Cutting Stock Problem}
Our work is intimately related to  the classical work of \cite{cuttingstock} on the cutting stock problem which originated in the paper industry.  In  \cite{cuttingstock} the authors determine how to satisfy a set of demands for rolls of paper of various widths given rolls of paper of longer widths while minimizing the scrap.  In this line of work there are a massive number of variables where each variable corresponds to a different way to cut a roll.  Each way of cutting a roll is called a pattern.  The number of possible patterns can grow exponentially in the number of unique widths demanded.  

The following very small scale example illustrates the concept of a pattern.  Consider cutting a 100 meter wide roll of paper in the context of demands for 30 and 45 meter long rolls.  Two example patters would be (1) cut the roll into  three 30 meter long rolls with 10 meters of scrap (2) cut the roll into one 45 meter long roll and one 30 meter long roll with 25 meters of scrap.   Optimization is formulated as an integer linear program (ILP) that is NP-Hard.  The value of a variable in a solution to the ILP denotes the number of rolls that are cut using the pattern associated with that variable.   The integer linear program is relaxed to a linear program allowing for a fractional number of each pattern to be used.  Optimization is initialized with a fixed number of patterns sufficient to satisfy all orders but likely with excess waste.  Optimization proceeds by solving the linear program given the current set of  patterns, followed by generating new patterns.  This is repeated until convergence. 

Finding the optimal pattern to add corresponds to solving a knapsack problem\cite{kellerer2004introduction,karp1972reducibility}.  The knapsack problem is NP-Hard however it can be approximated very well using a  dynamic programming based polynomial time approximation scheme.   This can be understood as maximizing the reward obtained by cutting a single roll by choosing a single pattern given reward values associated with each width of paper.  Here the rewards correspond to the value of the dual variables in the dual linear program.  In this manner the problem of optimizing over a massive number of patterns is reduced to finding a single pattern.

  In our applications we approximate high rank tensors with low rank tensors.  When generating new rank-1 tensors to add to our set of primitives  (working set) we construct a tensor to maximize sum of the element-wise products between it and another tensor corresponding to the dual variables.  In this manner the problem of optimizing over an infinite number of rank-1 tensors is reduced to iteratively fitting a rank-1 tensor. Like the knapsack problem, fitting the optimal rank-1 tensor is known to be NP-Hard \cite{mosttensorhard} however like the knapsack problem well studied  optimization schemes exist \cite{konda}.

 Producing integer solutions in the cutting stock may be done heuristically by greedily rounding up fractional values to integer values or by a principled but expensive branch and price algorithm\cite{barnprice}.  Fortunately our use of L1 regularization does not require integer solutions so branch and price is not needed.  Variants of the cutting stock problem include enforcing a hard constraint on the maximum number of patterns used and this can be interpreted as L0 regularization and is akin to our L1 regularization.
 
   \subsubsection{Marginals in Graphical Models}
Our work is related to the work of \cite{belanger2013marginal} (which is extended in \cite{krishnan2015barrier}). While column generation is not explicitly used or mentioned, their approach is very similar to column generation and their problem has a cross entropy loss like in Family Two.  In this paper the authors attack the problem of computing the marginal distributions of variables in a Markov random field (MRF).  This is done by minimizing standard Bethe-style convex variational objective.  The authors introduce an algorithm to construct a probability distribution over solutions and adds solutions to the probability distribution greedily.  At each step of the algorithm a new solution is generated and added to the probability distribution.

  Choosing the optimal solution to add corresponds to MAP inference which is an integer program and is often NP-Hard.  The potentials for MAP inference are a function of the current probability distribution and the potentials that define the original MRF, unlike the dual variables in our methods.  MAP Inference can be challenging because problem structure such as sub-modularity \cite{alphagraph} which allows for efficient MAP inference is lost when present in the original MRF however LP relaxation techniques\cite{kolmogorov2006convergent,sontag} can be used to produce efficient approximate solutions.  Once computed a line search is used to determine how much to weight this solution versus the other solutions.  The determination of the optimal solution via MAP inference can be interpreted as adding a primal variable (primitive) to the working set while the line search  is a greedy move towards optimizing the cross entropy loss.

\subsubsection{Generating Multiple Primal Variables }
Finding the optimal primitive variable to add in our applications is NP-Hard but multiple local optima can be generated concurrently given the same set of dual variables with each optima being computed on a separate CPU.  This can be contrasted to the work of \cite{yarkonytracker} where large numbers of variables each of which improves the objective are generated in polynomial time concurrently as part of a common operation.  In \cite{yarkonytracker}, tracking thousands of objects in video is formulated as an integer program which is relaxed to a linear program where each variable corresponds to an entire track of an object.  Finding the optimal variable to add to the working set corresponds to solving a dynamic program which is polynomial time solvable to global optimality.  However dynamic programming provides not only the optimal solution but the optimal solution passing through each position in space-time for free.  Each of these or a subset of the most violated can be added to the working set.  \cite{yarkonytracker} adds one thousand such tracks per iteration allowing for very fast inference.  Surprisingly solving the LP is not hindered by the addition of large numbers of tracks by huge working sets.  In our approach computing multiple local optima allows for some of the benefits of multiple solutions per dual solution it does not take advantage of the scale for easy generation of variables of \cite{yarkonytracker}.   %it does not scale in the same 

\subsection{ Low Rank Tensor Decompositions}

\subsubsection{Orthogonal Tensor Decompositions}
 In \cite{animatensor}, the authors attack the problem of latent variable modeling leveraging  modeling assumptions about how the data was generated.  If these assumptions are obeyed then they can guarantee globally optimal recovery of the data generating model.  Inference consists of computing a low rank approximation of an orthogonal tensor generated using the triple co-occurrence of moments in the data set (plus whitening).  The work of \cite{animatensor} is associated with powerful statistical guarantees.  However if the modeling assumptions are not obeyed by the data generation process the recovered solution which is intended to describe a mixture probability of distributions need not produce probability distributions though these can be rounded to probability distributions to produce approximate solutions.
 
 In contrast to the work of \cite{animatensor}  we focus on optimization, and we ensure that the outputted representation follows the rules that define the primitives.   Like \cite{animatensor} we use the power iteration to generate primitives though only in the context of creating a rank-1 representations of the dual variables.  
   
   %  We do not study the statistical efficiency of our approach in this paper and instead focus on minimizing loss functions. 

\subsection{Gradient Descent and Alternating Least Squares}

Two common and powerful approaches for tensor decomposition are gradient descent \cite{acar2008cpopt} and alternating least squares  (ALS)\cite{carroll1970analysis,konda,harshman1970foundations}. Gradient decent optimizes all modes of the tensor concurrently and converges to a local optima of the objective.  Alternatively ALS  solves for one of the modes of the tensor at a time keeping the other modes fixed.  This is achieved via least squares and  is coordinate-wise optimal.  ALS is not guaranteed to achieve global optimality but in practice is highly successful and benefits from the lack of a need for a step size as in gradient methods\cite{konda}.   

In Appendix \ref{nonsym1}, \ref{nonsym2} we show coordinate-wise updates for Family One and Family Two respectively while gradient descent updates are present in Section \ref{tensec1}, and in Appendix \ref{nonsym1}.  Since we compute only a single, rank-1 tensor at a time our method does not solve least squares problems during coordinate-wise updates.  Instead simpler closed form updates exist; for Family One this involve computing a gradient and obtaining an exact solution for the mode. 

A major difference between the gradient descent approaches, and ALS in contrast to our work is the form of regularization.  For gradient descent and ALS approaches the number of rank-1 terms is fixed in the beginning of optimization while in our approach it is not.  Thus gradient  descent and ALS  have implicit L0 regularization.  We have no ability to enforce an L0 norm  in our method and must be content to approximate it with an L1 norm.  

Another point of difference is that our approach generates new rank-1 tensors but does not adapt those that have already been produced.  Gradient descent and ALS continuously alter the tensors they are operating on while ours adds tensors to the working set during optimization leaving the previous ones fixed.   Our approach can be used in complementary way to ALS and gradient descent so as to update the rank-1 tensors in the working set perhaps speeding up inference.

\subsubsection{Constrained Tensor Decomposition via the Alternating Direction of Multipliers Method}
In  \cite{liavas2015parallel} the authors employ the  alternating direction of multipliers method \cite{boyd2011distributed} (ADMM) to break down constrained tensor factorization problems into separate problems (sub-problems) that are enforced to have a common solution using Lagrange multipliers.  Solutions to the sub-problems are much easier to compute than solving the original constrained problem though they must be re-solved many times.  This difficulty can be reduced by leveraging parallel computation.  

Lagrange multipliers are used to couple the problems together and operate in such a way as to not make the unconstrained problems more difficult to solve.  The work of \cite{liavas2015parallel} is extended in \cite{huang2015flexible} where it is demonstrated to produce state of the art results for non-negative tensor factorization, and other problems. Optimization is guaranteed to converge to a stationary point of the constrained objective.  Our approach can be used in a complementary way to \cite{liavas2015parallel} by using ADMM during the stage of generating new tensors. However unlike in \cite{liavas2015parallel,huang2015flexible} only a single rank-1 tensor would be constructed at a time.  

\section{Conclusions and Future Work }
\label{conc}

In this document we  apply column generation to approximating complex structured objects via a set of primitive structured objects under two families of loss functions with L1 regularization encouraging the use of few structured primitive objects.  We attack approximation using convex optimization over an infinite number of variables each of which are generated on demand using the corresponding dual problem.  We apply our approach to producing low rank approximations to large 3-way tensors.  Our work provides a broad domain for extensions and we note a few below.  

\textbf{Increasing Scale:}  Solving much larger problems will require clever use of LP and QP solvers.  One can attack this by using sub-optimal dual solutions to compute violated constraints.  Similarly one can use the LP and QP solvers in ways that do not restart optimization from scratch each time called or which forget constraints that are not active yet slow optimization. 

\textbf{Diverse Solutions: }Additional speed may be obtained by applying sampling and deterministic sampling approaches such as \cite{papandreou2011perturb,welling2009herding,batra2012diverse,krishnan2015barrier} to add diverse columns to $\hat{M}$ after each solution to the LP or QP.  This would be useful in a domain where solving the  QP/LP is significantly more time intensive than maximizing $\theta^tm$. 

 \textbf{Size Reduction:} Applications of tensor size reduction methods as used in \cite{animatensor} may prove to be invaluable to extending our work to much larger problems especially if the corresponding statistical guarantees can be preserved. This is challenging as the constraints on elements $m$ change in the new reduced size space.  

%In addition leveraging, and extending the guarantees of  \cite{animatensor} is a fertile area for research.
%Another line of work is to apply column generation methods for computing marginals in graphical models by extending the work of \cite{belanger2013marginal,krishnan2015barrier}.

\textbf{L0 regularization: }  The application of branch and bound/price techniques \cite{barnprice} after the construction of $\hat{M}$ using our approach with L1 regularization, may prove useful for optimizing under L0 regularization. 

\textbf{Priors and  Mixture Modeling:  } We did not explore the use of priors over the variance of gaussians for learning mixture of gaussians and optimization over the variance.  We suspect this can be accomplished by  altering the L1 regularizer to penalize lower variance gaussians and enforcing this via the conjugate prior for the gaussian, which is the normal Wishart prior \cite{conjpriowish}.  This may be useful in the domain of quantum chemistry.  

\textbf{Sparse Coding/Topic Models} Applying the column generation approach to the Family One problem of sparse coding requires introducing a term to limit the total number of primitives used across documents (data samples).  This can be done in an L1 sense by associating each primitive with variable in the QP that is associated with the maximum amount this primitive is used by any document.  L1 regularization is then applied to these variables in the QP with the aim of limiting the total number of primitives used across documents.  A separate L1 regularization is used to encourage each document to have a sparse representation.  The Family Two version of this would be a topic model.  %however making such practical would appear to be rather difficult

\textbf{Deep Relationships: }  We suspect that there is a relationship explaining the effectiveness of large single hidden layer neural networks studied in \cite{caruanamodel}, using our work and boosting \cite{freund1999short}.

{\small
\bibliographystyle{ieee}
\bibliography{BibClean}
}

\appendix

\section{Family One: Dual Derivation}
 \label{fam1dualderiv}

We now derive the dual problem of Family One given the primal problem of Family One.  We begin with the Family One objective function after the introduction of the surrogate complex structured object $K$.   
  \begin{align}
 % \label{fam2primalbef}
 \min_{\substack{w \geq 0 \\ K \geq 0}}\frac{1}{2} K^t K+\vec{\ell}^tw\\
\nonumber \mbox{s.t. }K \geq T-Mw\\
\nonumber K \geq Mw-T
 \end{align}
 
 Next we replace the constraints with Lagrange Multipliers.  
 
\begin{align}
\mbox{Eq }\ref{fam2primalbef}=\min_{\substack{w \geq 0\\K\geq 0}}\max_{\substack{\lambda \geq 0\\ \psi \geq 0}} \frac{1}{2}K^tK+\vec{\ell}^t w\\
\nonumber+\psi^{t}  (-T+Mw-K)\\
\nonumber + \lambda^{t} ( T-Mw-K)
\end{align}
Since Eq \ref{fam2primalbef} is a convex quadratic program we are able to flip the order of the  $\min$ and the $\max$ without altering the value of the objective.  
\begin{align}
\label{dervpoint}
\mbox{Eq }\ref{fam2primalbef}=\max_{\substack{\lambda \geq 0\\ \psi \geq 0}} \min_{\substack{w \geq 0  \\ K \geq 0}} \frac{1}{2}K^tK+\vec{\ell}^t w\\
\nonumber+\psi^{t}  (-T+Mw-K)\\
\nonumber + \lambda^{t} ( T-Mw-K)
\end{align}

We now take the first derivative of Eq \ref{dervpoint}  with respect to $K$ and then solve for $K$ in terms of the dual variables $\lambda$ and $\psi$.  
\begin{align}
\label{redsolve}
0=K-\psi-\lambda\\
\nonumber K=\psi+\lambda
\end{align}

Given the closed form solution for $K$  in Eq \ref{redsolve} we write Eq \ref{fam2primalbef} with $K$ replaced by $\psi+\lambda$.

\begin{align}
\mbox{Eq }\ref{fam2primalbef}=\max_{\substack{\lambda \geq 0\\ \psi \geq 0}} \min_{\substack{w \geq 0}} \frac{1}{2}(\lambda+\psi)^t(\lambda+\psi)+\vec{\ell}^t  w\\
\nonumber+\psi^{t}  (-T+Mw-\lambda-\psi)\\
\nonumber + \lambda^{t} ( T-Mw-\lambda-\psi)
\end{align}

We now group the terms by primal variable.  

\begin{align}
\mbox{Eq }\ref{fam2primalbef}=\max_{\substack{\lambda \geq 0\\ \psi \geq 0}} \min_{\substack{w \geq 0}} \frac{-1}{2}(\lambda+\psi)^t(\lambda+\psi)+\lambda^t T-\psi^t T\\
\nonumber+(\vec{\ell}-\lambda^tM+\psi^tM)w
\end{align}

We now convert the Lagrangian to a convex quadratic program by converting the primal variables $w$ into constraints. 

\begin{align}
\mbox{Eq }\ref{fam2primalbef}=\max_{\substack{\lambda \geq 0\\ \psi \geq 0}} \frac{-\lambda^t\lambda}{2} -\psi^t\lambda -\frac{\psi^t\psi}{2}+T^t \lambda- T^t\psi\\
\nonumber \vec{\ell}\geq M^t\lambda-M^t \psi
\end{align}

\section{Family Two: Dual Derivation}
 \label{fam2dualderiv}
 
We now derive the dual problem of Family Two given the primal problem of Family Two.  We begin with the Family Two objective function after the introduction of the surrogate complex structured object $K$ and the replacement of $\log$ by a concave envelope of affine functions.  

  \begin{align}
  \label{fam2expandz}
\min_{\substack{w \geq 0 \\ 1^t w \leq 1\\ K \geq 0}} T^tK+\vec{\ell}_a^tw\\
\nonumber \mbox{s.t. }-K \leq  \log(1 \eta) +\frac{(Mw-1\eta)}{\eta} \quad  \forall \eta \in (0,1]
 \end{align}

We now insert Lagrange multipliers $\alpha$ and $\beta$.  
\begin{align}
\mbox{Eq } \ref{fam2expandz}=\min_{\substack{w \geq 0\\ K \geq 0}}\max_{\substack{\alpha \geq 0\\ \beta \geq 0}}T^tK+\vec{\ell}_a^t w\\
\nonumber \alpha(1^tw-1) \\
\nonumber \sum_{\eta \in (0,1]}\beta^{t\eta}(-K- \frac{1}{\eta}Mw+1-1\log(\eta))	%\quad \quad \forall u\in \mathcal{U}
\end{align}

We now group by primal variable.  

\begin{align}
\mbox{Eq } \ref{fam2expandz}= \min_{\substack{w \geq 0\\ K \geq 0}}\max_{\substack{\alpha \geq 0\\ \beta \geq 0}} \sum_{\eta \in (0,1]}\beta^{t\eta}(1-1\log(\eta))-\alpha\\
\nonumber (T^t-\sum_{\eta \in (0,1]}\beta^{t\eta})K\\
 \nonumber (\vec{\ell}_a^t+\alpha1^t-\sum_{\eta \in (0,1]} \frac{1}{\eta} \beta^{t\eta}M)w
\end{align}

We now convert primal variables to hard constraints. 

\begin{align}
\mbox{Eq } \ref{fam2expandz}= \max_{\substack{\alpha \geq 0\\ \beta \geq 0}} \sum_{\eta \in (0, 1]}\beta^{t\eta}(1-1\log(\eta))-\alpha\\
\nonumber T^t-\sum_{\eta \in (0, 1]} \beta^{t\eta} \geq 0\\
\nonumber \vec{\ell}_a^t+\alpha1^t-\sum_{\eta \in (0, 1]} \frac{1}{\eta} \beta^{t\eta}M \geq 0
\end{align}
We now take the transpose of our objective and move terms.  

 \begin{align}
\mbox{Eq } \ref{fam2expandz}= \max_{\substack{\alpha \geq 0\\ \beta \geq 0}} -\alpha +\sum_{\eta \in (0,1]}(1-1\log(\eta))^t\beta^{\eta}\\
\nonumber T \geq \sum_{\eta \in (0,1]}\beta^{\eta}\\
\nonumber \vec{\ell}_a+1\alpha \geq M^{t} \sum_{\eta \in (0,1]} \frac{1}{\eta} \beta^{\eta}
\end{align}

\section{Optimizing $\theta^t m$, Family One:  Non-Symmetric Tensors}
\label{nonsym1}
We now study $\max_{m \in M}\theta^t m$  in the domain where $m$ corresponds to fitting a non-symmetric 3-way tensor.  Thus each column of $M$ also corresponds to a non-symmetric 3-way tensor.  We describe $m$ using vectors unit vectors  $v^a,v^b,v^c$ which are indexed by $i,j,k$ respectively.  The non-vectorized form of $m$ is denoted $\bar{m}$ and defined below.   %$i,j,k \in \{0,1,2...N-1\}$.  We define $\bar{m}$ in terms of $v$ below.

\begin{align}
\bar{m}_{ijk}=v^a_i v^b_j v^c_k
\end{align}

We write optimization below. 
%.  Thus $v\geq 0$ and $1^tv = 1$
\begin{align}
\max_{m \in M}\theta^t m=\max_{\substack{v^a,\; v^b \; v^c \\ v^{at}v^a = 1 \; v^{bt}v^b = 1\\  v^{ct}v^c = 1 } }\sum_{ijk}\theta_{ijk}v^a_i v^b_j v^c_k
\end{align}

The projected gradient update for $v^a_i$ for all $i$ is written below.

\begin{align}
\dot{v}^a_i \leftarrow v^a_i+\mbox{stepsize} \sum_{jk}\theta_{ijk}v^b_j v^c_k 	\quad \forall i\\
\nonumber v^a \leftarrow \frac{\dot{v}^a}{\dot{v}^{at}\dot{v}^a}
\end{align}

As an alternative or supplement to gradient descent one can perform coordinate-wise updates which update one of the vectors given the other vectors.  Consider optimizing $v^a$ given $v^b$ and $v^c$.  Observe that gradient of $\sum_{ijk}\theta_{ijk}v^a_i v^b_j v^c_k$ with respect to $v^a$ has no dependency on $v^a$ and $v^a$ is restricted to be unit norm.  Thus the  optimal $v^a$ is thus simply the gradient of $\sum_{ijk}\theta_{ijk}v^a_i v^b_j v^c_k$ with respect to $v^a$ properly scaled to be a unit vector.  Thus we can write the update for $v^a$ as the following expression.  

\begin{align}
\dot{v}^a_i\leftarrow \sum_{jk}\theta_{ijk}v^b_j v^c_k 	\quad \forall i\\
\nonumber v^a \leftarrow \frac{\dot{v}^a}{\dot{v}^{at}\dot{v}^a}
\end{align}

Optimization over $v^a,v^b,v^c$ proceeds by cycling through $a$,$b$,$c$ optimizing $v^a$, then $v^b$, then $v^c$.  Higher order generalizations of the singular value decomposition may find use here allowing for example $v^a$,$v^b$ to be updated given $v^c$.

\section{Optimizing $\theta^t m$, Family Two:  Non-Symmetric Tensors}
\label{nonsym2}
We now consider updates for 3-way non-symmetric tensors in Family Two using the notation of Section \ref{nonsym1}.  Recall that  $\theta$ is non-negative and that we normalize $\theta$ to sum to one and shift the normalization constant outside of the $\max$. We now write optimization over the $\log$ of $\max_{m \in M}\theta^t m$.
  \begin{align}
  \label{befzNS}
 \max_{\substack{v^a\geq 0, \; v^b\geq 0,\; v^c\geq 0 \\ 1^tv^a = 1, \; 1^tv^b=1,\; 1^tv^c=1}} \log(\sum_{ijk} \theta_{ijk}v^a_i v^b_j v^c_k)
 \end{align}
 
 As in EM methods for inference in probabilistic models we define a proposal probability distribution $z$ indexed by $ijk$.  We initialize $z$ as follows to reflect the probability distribution $\theta$ though this initialization is heuristic and random initialization of $z$ is also valid.  
 \begin{align}
 z_{ijk}\leftarrow \frac{\theta_{ijk}}{\sum_{\dot{i}\dot{j}\dot{k}}\theta_{\dot{i}\dot{j}\dot{k}}}  %The 6 is to cover the symmetry in $\theta$.  
\end{align}
We now multiply and divide by $z$ as is standard in EM methods and apply Jenson's inequality .  
  \begin{align}
 \mbox{Eq } \ref{befzNS}= \max_{\substack{v^a\geq 0, \; v^b\geq 0,\; v^c\geq 0 \\ 1^tv^a = 1, \; 1^tv^b=1,\; 1^tv^c=1}} 
\log(\sum_{ijk} \frac{z_{ijk}}{z_{ijk}}\theta_{ijk}v^a_i v^b_j v^c_k)\\ %\; \; \forall[ v\geq0; 1^tv=1]\\
\nonumber \geq \max_{\substack{v^a\geq 0, \; v^b\geq 0,\; v^c\geq 0 \\ 1^tv^a = 1, \; 1^tv^b=1,\; 1^tv^c=1}} 
 \sum_{ijk}-z_{ijk}\log z_{ijk}+z_{ijk}\log \theta_{ijk} \\
\nonumber + \sum_{ijk}z_{ijk}\log v^a_i+\sum_{ijk}z_{ijk}\log v^b_j+\sum_{ijk}z_{ijk}\log v^c_k
 \end{align}

We now add a  Lagrange multiplier $\tau^a,\tau^a,\tau^c,$ to enforce that $v^a,v^b,v^c$ sum to one respectively.  There will be no need to enforce non-negativity in optimization.  

  \begin{align}
 \max_{\substack{v^a\geq 0\\ v^b\geq 0\\  v^c\geq 0} }\min_{\substack{\tau^a \in (-\infty,\infty)\\ \tau^b \in (-\infty,\infty)\\ \tau^c \in (-\infty,\infty)}}
 \tau^a(1-1^tv^a)+ \tau^b(1-1^tv^b)+ \tau^c(1-1^tv^c)+\sum_{ijk}-z_{ijk}\log z_{ijk}\\
  \nonumber +z_{ijk}\log \theta_{ijk}+ \sum_{ijk}z_{ijk}\log v^a_i  +\sum_{ijk}z_{ijk}\log v^b_j+\sum_{ijk}z_{ijk}\log v^c_k
\end{align}

We now write optimization with respect to $v^a_i$.  We now take derivative with respect to $v^a_i$ and set it equal to 0.  

\begin{align}
0=-\tau^a+\sum_{jk} (z_{ijk})\frac{1}{v^a_i}\\
\nonumber \tau v_i=\sum_{jk} z_{ijk}\\
 \nonumber v^a_i=\frac{1}{\tau^a}\sum_{jk} z_{ijk}
 \end{align}

Observe that $v^a_i\propto \sum_{jk} z_{ijk}$.  Since it is the case that  $1^t v^a=1$.  Then the following is true.  
\begin{align}
\tau^a=\sum_{ijk} z_{ijk}
\end{align}

The optimizing updates for $z_{ijk}$ set $z_{ijk}$ proportional to $\theta_{ijk}v^a_i v^b_j v^c_k$ based on the standard application of the tightest bound for Jenson's inequality.   Updates for $v^b_j$ and $v^c_k$ follow the same form as $v^a_i$.  Therefore the final updates are as follows.  
\begin{align}
\label{zvupdatesNS}
z_{ijk} \propto \theta_{ijk}v^a_i v^b_j v^c_k\\
\nonumber v^a_i\propto \sum_{jk} z_{ijk}\\
\nonumber v^b_j\propto \sum_{ik} z_{ijk}\\
\nonumber v^c_k\propto \sum_{ij} z_{ijk}
\end{align}
We repeatedly update $z$ then $v$ until convergence.  

%\subsection{possible}
%Singular value decomposition  may find use here allowing for example $v^a$,$v^b$ to be updated given $v^c$.  
%Observe that $v^b $ and $v^c$ can be optimized in an identical way.  

%\begin{align}
%\min_{v^a,v^b}\sum_{ijk}\theta_{ijk}v^a_i v^b_j v^c_k\\
%=\min_{v^a,v^b}\sum_{ij}v^a_i v^b_j (\sum_k v^c_k\theta_{ijk})\\
%\min_{v^a,v^b}\sum_{ij}v^a_i v^b_j \theta^{ab}_{ij}\\
%=v^{at}\theta^{ab}v^b
%\end{align}

%Observe the following
%\begin{align}
%\theta-
%\end{align}

\end{document}